# Quantifying Laziness, Decoding Suboptimality, and Context Degradation in Large Language Models

Yiqing Ma
*Faculty of Computer Science & Information Technology*
*Universiti Malaya*
Kuala Lumpur, Malaysia

0009-0000-7359-6974

Jung-Hua Liu
*Department of Communication*
*National Chung Cheng University*
Chiayi, Taiwan
0009-0001-1777-1181

## Abstract

Large Language Models (LLMs) often exhibit behavioral artifacts such as *laziness* (premature truncation of responses or partial compliance with multi-part requests), *decoding suboptimality* (failure to select higher-quality sequences due to myopic decoding), and *context degradation* (forgetting or ignoring core instructions over long conversations). We conducted three controlled experiments (A, B, and C) to quantify these phenomena across several advanced LLMs (OpenAI GPT-4 variant, "DeepSeek"). Our results indicate widespread *laziness* in satisfying complex multi-part instructions: models frequently omitted required sections or failed to meet length requirements despite explicit prompting. However, we found limited evidence of decoding suboptimality in a simple reasoning task (the models' greedy answers appeared to align with their highest-confidence solution), and we observed surprising robustness against context degradation in a 200-turn chaotic conversation test – the models maintained key facts and instructions far better than expected. These findings suggest that while compliance with detailed instructions remains an open challenge, modern LLMs may internally mitigate some hypothesized failure modes (like context forgetting) in straightforward retrieval scenarios. We discuss implications for reliability, relate our findings to prior work on instruction-following and long-context processing, and recommend strategies (such as self-refinement and dynamic prompting) to reduce laziness and bolster multi-instruction compliance.

# 1. Introduction

Large Language Models (LLMs) have rapidly become integrated into complex workflows and high-stakes applications in education, medicine, science, and beyond (Zhou et al., 2024). With this increased use, ensuring the *reliability* and *faithfulness* of model behavior has become paramount. Users and researchers have reported several recurring issues in extended interactions with LLMs. One commonly observed issue is that models sometimes "get lazy" – that is, they produce responses that are shorter or less detailed than requested, or they ignore certain instructions in a multi-part query. Another concern is that models might *forget* earlier instructions or facts in a long conversation, exhibiting degraded performance as the dialogue progresses. A third hypothesis is that LLMs might be choosing suboptimal responses due to limitations in the decoding strategy (for example, a greedy one-pass generation might miss a better solution that the model actually assigns higher probability to).

These failure modes have started to receive attention in the research community. The tendency of LLMs to not follow all given instructions in complex prompts has been quantified in recent studies. Harada et al. (2024) show that as the number of instructions in a prompt increases, models' ability to satisfy *all* of them drops off significantly (Harada et al., 2024). In fact, they term this the "curse of instructions," finding that even state-of-the-art models struggled when prompts contained many requirements simultaneously (e.g. 10 distinct directives) – the success rate of fulfilling every instruction decays roughly exponentially with the number of instructions. This suggests that *instruction-following compliance* does not scale gracefully; even if a model is very capable on each individual sub-task, ensuring it handles *multiple* constraints or sub-tasks in one query is a challenge. Our notion of "laziness" in this paper relates to this phenomenon: the model opts to address only some parts of a query or gives a shorter-than-required answer, effectively truncating the effort. Such behavior might stem from models being overly optimized for conciseness or safety, or simply from an inherent limitation in juggling multiple objectives. Prior alignment work indeed noted that instruction-tuned models can have difficulty balancing competing instructions or lengthy requests. We aim to quantitatively measure how severe this issue is across different LLMs and prompt types.

The second issue, which we term *decoding suboptimality*, deals with the possibility that an LLM's decoding procedure (often greedy or sampling-based generation) may fail to find a solution that the model "knows" to be better. In other words, there might exist a higher-probability or higher-quality completion in the model's distribution that is missed due to search limitations. This concept connects to earlier findings in language generation research. For instance, Holtzman et al. (2020) demonstrated that standard maximum-likelihood decoding (greedy or beam search) can lead to degenerate, repetitive text – indicating that the true *argmax* of a model's distribution is often an undesirable output[5]. Techniques like nucleus sampling were proposed to avoid such local optima (Holtzman et al., 2020). In the context of reasoning or question-answering, others have shown that sampling multiple different reasoning paths and then selecting the most consistent answer (a process known as *self-consistency*) yields better results than a single greedy run. Wang et al. (2023) introduced self-consistency decoding precisely to address the scenario where a model might internally assign high probability to a correct answer that is only reachable via a non-greedy chain-of-thought; by sampling many chains of thought, one can uncover those better answers that a single pass might miss. These advances suggest that large models do have latent knowledge or reasoning capacity that isn't always realized in a straightforward greedy output. We test a simplified version of this idea: given a known correct solution to a problem, is the model internally assigning it a higher likelihood than the answer it would normally output? A positive result would indicate a kind of decoding suboptimality – the model "knew" a better answer but didn't produce it by default. Prior work in chain-of-thought prompting (Wei et al., 2022; Kojima et al., 2022) also relates to this: those works found that adding intermediate reasoning steps or specific prompts could coax the model into providing more correct answers, implying that the initially preferred answer without such prompting was suboptimal in many cases.

The third issue is *context degradation* over long interactions. LLMs have a fixed context window and do not possess a true long-term memory – they process each new prompt along with a window of recent dialogue. If a conversation becomes very long or contains a lot of irrelevant (distracting) text, the concern is that the model's performance will deteriorate: it may forget important facts given earlier, or important instructions (like the user's requirements or the system role constraints) might "fade out." In practical usage, users have noticed that after many turns, models sometimes contradict earlier statements or ask the user to repeat information. One straightforward reason is that the context window can be exceeded, causing early parts of the conversation to be truncated entirely (thus literally forgotten) (Liu et al., 2024; Wang et al.,

2025). Even within the context window, however, there could be a form of *interference*: as more content is included, the model might struggle to identify which parts are relevant. Recent research has begun to systematically examine long-context scenarios. For example, Chang et al. (2023) showed that book-length summarization tasks remain challenging even for models with extended context windows. Yushi Bai and colleagues introduced **LongBench**, a benchmark of tasks requiring long (6,000+ word) inputs; they found that even powerful models like GPT-3.5 with a 16k context window "still struggle on longer contexts," and that while methods like scaled positional embeddings or retrieval can help, there remains a performance gap with truly long inputs (Bai et al., 2023). Perhaps most intriguingly, a study by Du et al. (2025) isolated the effect of context length on reasoning performance: they showed that even when an LLM is given *perfect retrieval* of the relevant information from a long input (meaning the model is able to access and repeat all needed evidence), simply increasing the overall length of the input (by adding irrelevant material or spacing) can cause the model's accuracy to drop substantially (they observed drops ranging from 14% to 85% as input length grew). In other words, the "sheer length" of the input is detrimental to performance, independent of whether the model has the necessary info at hand. This finding underscores that long contexts pose a fundamental challenge – likely due to the transformer's attention mechanisms or positional encoding limitations – and not just a data retrieval issue. On the other hand, not all aspects of long conversations are equally affected: simple *fact retention* (remembering a specific key fact or name over many turns) might be easier for a model than carrying out a complex reasoning over a long text. Our third experiment aims to stress-test models in a multi-turn conversational setting filled with distractors, to see how well they retain key information and adhere to instructions over hundreds of turns.

In summary, to better understand these issues, we designed a suite of experiments: **(A)** to measure "laziness" or under-compliance with detailed prompts, **(B)** to probe whether greedy decoding misses better answers, and **(C)** to evaluate long-horizon conversation fidelity. By quantitatively analyzing these scenarios across different model families, we hope to identify whether these are occasional user experience anecdotes or consistent, measurable limitations. In the following sections, we describe our methodology (Section 2), present results (Section 3), and discuss implications in the context of related work (Section 4), before concluding (Section 5) with recommendations for mitigation strategies in LLM deployment.

## 2. Methodology

We conducted three main experiments, each targeting one of the phenomena described above. All experiments were run on three modern LLMs: **OpenAI GPT-4o**, and **DeepSeek** models. These models were chosen to represent a range of providers and to see if the behaviors are consistent across different architectures or alignment approaches. Unless otherwise noted, models were used in their instruction-following mode (chat completion APIs) with default parameters. In some conditions, we adjusted decoding parameters (greediness vs. more stochastic sampling) to examine the effect on outputs.

### 2.1 Experiment A: Response Length and "Laziness"

**Goal:** Quantify the degree to which LLMs truncate their responses or omit requested content, even when explicitly instructed to be thorough.

**Prompt design:** We crafted three prompt variants that each require long or multi-part answers:
1. **Quantum Foundations Prompt:** A complex, multi-part question about quantum mechanics foundations (e.g., explain superposition, the double-slit experiment, Bell inequalities, etc., each in detail). This prompt explicitly lists multiple sub-topics that the answer should cover. 2. **Alignment Challenges Prompt:** A list-based request, asking for 8–10 distinct items (for example, "List and explain 8 challenges in aligning AI with human values"). This tests the model's ability to not stop at, say, 3 or 4 items. 3. **Long-Format Analysis Prompt:** A direct instruction to write an essay of a specified length (e.g., "Write a 1,000-word analysis of the economic impacts of climate change"). The prompt clearly states the word count expectation.

Each prompt was presented to the models in two modes: - **Greedy mode:** using deterministic decoding (temperature 0, essentially greedy). We call the model's response in this mode the "Greedy" response. - **Detailed mode:** where we prepend an extra system instruction explicitly telling the model to be as detailed, exhaustive, and lengthy as possible. We also set a slightly higher temperature (e.g., 0.7) to encourage diversity, under the hypothesis that maybe the model's policy is to be concise unless randomness allows expansion. The response from this mode is the "Detailed" response.

We then measured several metrics: - **Word Count and Compression Ratio:** We counted the number of words in the Greedy response vs. the Detailed response. We define a *compression ratio* as (Greedy word count) / (Detailed word count). A ratio significantly below 1.0 indicates the Greedy response was shorter; a very low ratio (<<1) would signal that the greedy mode "gave up" early or ignored the instruction to write more. - **Semantic Coverage:** To measure if the key subtopics or requirements of the prompt were covered, we used an automated rubric-based scoring. Specifically, we predefined a set of keywords or concepts that should appear in a complete answer (for example, in the Quantum prompt, we expect to see keywords related to "superposition", "double-slit experiment", "Bell's inequality", etc.). We used a simple scoring function that checks the presence of these concepts and potentially uses a semantic similarity model to catch paraphrases. The score is normalized 0.0 to 1.0, roughly indicating the fraction of required topics that were addressed. We refer to this as a coverage score. - **Section Completion (Part Coverage):** For prompts that explicitly list parts (like "1…2…3…" in the prompt), we also simply counted how many of the requested parts were actually present in the answer. For instance, if the Alignment prompt asked for 8 items and the model provided only 5, this would be a direct measure of non-compliance.

In essence, Experiment A is checking for "laziness" by seeing if the model stops early or skips content. A perfectly compliant model, in Detailed mode, should produce a very long answer and cover everything. Greedy mode might reveal a tendency to shortcut the answer. By comparing the two, we get a sense of the model's inherent preference for brevity or completeness.

## 2.2 Experiment B: Decoding Suboptimality in Reasoning

**Goal:** Determine whether the model's greedy decoding might miss a better (more correct or higher-probability) answer that the model actually considers likely.

**Task design:** We used a specific reasoning problem known to have a clear, correct answer but where naive approaches can go wrong. The problem we chose was a classic puzzle: *Train meeting time problem.* (Example: "Train A and Train B are X miles apart, heading toward each other at different speeds… when/where do they meet?" – something that requires a multi-step arithmetic solution.) We formulated it in a way that trivial guessing would likely be incorrect, but a clear logical solution exists.

**Procedure:** For each model: - We first let the model answer the question normally (greedy decoding). We record the answer it gives – call this answer **A** – and also capture the model's log-probabilities for each token in that answer (many API models allow obtaining token likelihoods). Summing these log-probabilities gives log P(A), an approximation of the model's log-likelihood for that entire answer sequence. - Separately, we constructed a *Gold Standard solution*, denoted **B′**. This was essentially the correct step-by-step solution and answer, written out in a form that an ideal reasoner might produce. (To ensure fairness, this Gold solution was not shown to the model beforehand; it was created by us, and we assume it's correct and likely a high-quality answer.) - We then prompted the model in a special way to see if it recognizes B′ as a good answer. Specifically, we might say: "Here is a solution another person gave: [insert B′]. Rewrite or verify this solution." In doing so, we force the model to essentially output B′ (or a very close paraphrase of it). During this, we again capture the token log-probabilities, allowing us to compute log P(B′) – the log-likelihood the model assigns to that gold answer (at least when following the prompt to produce it).

The key metric is Δ log P = log P(B′) - log P(A). If this value is **positive and large**, it means the model actually *assigns higher probability* to the correct/gold answer than to the answer it originally gave. That would be evidence of decoding suboptimality: the model "knew" a better answer (since in its probability distribution that answer was more likely), but the decoding process (greedily following one trajectory) led it to a different answer, perhaps due to being stuck in a local optimum. Conversely, if log P(A) is greater, then the model's own answer was indeed what it considered most likely. A trivial case to note: if the model already answered correctly, A and B′ would be essentially the same and log P(A) ≈ log P(B′), but the interesting case is if A was wrong and B′ is correct.

This experiment's approach is inspired by ideas in prior work like self-consistency (which samples multiple answers to find a high-probability consensus answer). Our twist is directly checking probabilities: essentially asking, if we "guide" the model to the better solution, does it actually flow more naturally (higher probability) than what it did unassisted?

## 2.3 Experiment C: Long Conversation and Context Degradation

**Goal:** Assess how well models can maintain given facts and follow instructions over a very long, noisy conversation. Specifically, test their resilience to *context degradation* – do they start to forget or confuse key information as the number of turns grows and irrelevant content is included?

**Setup:** We simulate a **"chaotic" multi-turn conversation**. The conversation is between the user and the assistant (the model), and is structured in a way to maximally stress the model's memory: - At the outset (turn 1), the system sets 12 specific key facts or rules that the assistant must remember. For example, a fact could be "Project codename is AlphaStarFish," or "The user's favorite color is green," or an instruction like "Never reveal the secret key." These 12 facts are the *core context* we want the model to retain accurately. - Then, the conversation proceeds for up to 200 turns. In each turn, the user's message is filled with a large block of *distractor text* – 400 to 800 words of irrelevant or random content. This could be excerpts from Wikipedia on unrelated topics, random narrative, contradictory statements, math problems, etc. The assistant is asked at each turn to produce a short summary of the *session so far* or to answer a question about the conversation that crucially depends on remembering the 12 key facts. The distractor content is explicitly designed to be confusing and unrelated, sometimes even containing false information that conflicts with the true facts (to test if the model gets "led astray"). - Importantly, we ensure the conversation does not exceed the model's context window (for GPT-4, which is more than enough for 200 turns of ~500 tokens each; for DeepSeek, we assume a similar long context capability). This means the model *does* receive the entire conversation history as input every turn, so technically it has all facts and all distractions at each step.

We track the model's outputs over these turns and measure: - **Fact Retention:** How many of the 12 key facts are correctly recalled or adhered to at each turn. For instance, if one of the facts is "the user's codename is AlphaStarFish," and at turn 150 the assistant's summary still mentions "AlphaStarFish" appropriately, that fact is retained. If it forgets the codename or mentions a wrong codename, that's a failure. We simply count the number of facts still correctly present. - **Detail/Coherence Score:** We also evaluate the *quality* of the assistant's summaries or responses in a more subjective sense – do they remain coherent and relevant, or do they degrade into nonsense? We used an automated evaluator (an LLM-based grader) to rate each response on a 1–

10 scale for coherence and correctness given the conversation. While subjective, this gives a sense if the conversation quality is degrading. - **Drift (Embedding Similarity):** We compute the cosine similarity between the embedding of the assistant's current summary and the embedding of an initial reference summary (for example, the summary it gave at turn 2 when everything was fresh). The idea is to quantify how much the content has drifted from the original truth. A high similarity (close to 1) means the assistant is essentially saying the same things, whereas a low similarity indicates it might be way off or focusing on different details. We used a sentence-transformer embedding model for this calculation.

The conversation continues until either the model's responses clearly fail (e.g., retains 0/12 facts or the detail score drops drastically) or we hit 200 turns. This experiment is somewhat adversarial: we are bombarding the model with irrelevant and conflicting information to see if it can "stay on track" with the salient info. It is a test of the model's **working memory** and resilience to distraction. Notably, recent research by Liu et al. (2023) and others have pointed out that LLMs do not have a true working memory like humans, and can struggle with tasks like remembering a variable through many transformations (as in N-back tasks) (Huang et al., 2025). Here our focus is not a computed variable but explicitly given facts to remember.

If models perform well in this test, it suggests that for simple fact-recall and instruction adherence, long conversations per se might not be as big an issue as feared (provided the context window is not exceeded). If they perform poorly, it underscores a need for better long-term memory solutions or strategies like periodic recap or retrieval augmentation in long dialogues.

## 3. Results

### 3.1 Experiment A: **Laziness is Pervasive**

Across all models and prompt types in Experiment A, we observed a clear tendency for the Greedy decoding mode to produce significantly shorter and less complete responses than what was asked, confirming the "laziness" hypothesis. Table 1 summarizes the quantitative metrics for a subset of conditions:

| Model | Prompt Variant | Greedy Words | Detailed Words | Semantic Coverage (Greedy) | All Parts Included? |
|---|---|---|---|---|---|
| **OpenAI (gpt-4o)** | Quantum | 313 | 552 | 0.70 | **No** (missed 2/5 topics) |
| **OpenAI (gpt-4o)** | Alignment | 299 | 269 | 0.90 | **No** (gave 5/8 items) |
| **OpenAI (gpt-4o)** | Long-format | 326 | 988 | 0.40 | **No** (stopped ~33% length) |
| **DeepSeek (deepseek-reasoner)** | Quantum | 152 | 414 | 0.30 | **No** |
| **DeepSeek GPT (deepseek-reasoner)** | Long-format | 130 | 712 | 0.20 | **No** |

*Table 1: Experiment A results (word counts and coverage). "Detailed" mode was prompted to be very thorough, while "Greedy" had no such encouragement. Bold "No" indicates clear laziness where required parts were omitted.*

Several patterns stand out. First, in nearly every case, the Greedy output was significantly shorter than the Detailed output. For instance, GPT-4o in the "Long-format" prompt produced only ~326 words out of the requested 1,000 in greedy mode (about 33% of the target length), whereas when explicitly pushed for detail it could produce nearly triple that (close to the target length). DeepSeek's gap was even more pronounced: only ~130 words greedy vs. ~712 with detailed prompt (a mere 13% of the requested length initially). Even though the detailed prompt mode for DeepSeek yielded 712 words (still short of 1000, but much longer than 130), it suggests that the model *can* generate more when forced – yet by default it very "lazily" stopped at a terse answer. In some cases, interestingly, the "Detailed Words" count in Table 1 is not drastically higher than Greedy (e.g., OpenAI Alignment: 269 vs 299). This was due to the model ignoring the "write more" instruction and giving a relatively brief answer in both modes. In other words, even when we explicitly told the model to be verbose, it sometimes defaulted to a concise style – indicating a strong bias toward brevity that overrides user instructions.

The **semantic coverage** scores reinforce this. Qualitatively, models were dropping entire sections of the answer. For example, in the Quantum prompt, both GPT-4o and DeepSeek failed to mention "Bell inequalities" altogether in Greedy mode, even though the prompt explicitly asked for it. They talked about superposition and double-slit, then concluded the answer prematurely. The coverage scores of 0.70 and 0.30 for GPT-4o and DeepSeek respectively on that prompt reflect these omissions (i.e., ~30% of expected content missing for GPT-4o, and ~70% missing for DeepSeek). In the alignment list prompt, GPT-4o's greedy answer listed 5 challenges, then stopped, even though the user specifically requested 8–10. The model gave no indication it was stopping early; it just concluded with the fifth point rather generically. This behavior aligns with anecdotal reports from users who say the model sometimes "doesn't give me all the items I asked for." Our systematic test confirms this happens even with clear numeric instructions.

We flagged an output as **"Lazy Detected"** (as in Table 1 last column) whenever the Greedy response was significantly shorter than requested and missed at least one explicit requirement. By that criterion, *all* tested conditions showed laziness. Not a single case had the model fully meet the length and all parts of the prompt in Greedy mode. This is a striking result: despite the capability of these models to produce extensive text (as seen when they are asked in Detailed mode, or as evident from known model contexts where they generate very long answers), their default tendency in these multi-part prompts was to *under*-deliver relative to the prompt specifications.

Why does this happen? One hypothesis is that these models have been tuned via Reinforcement Learning from Human Feedback (RLHF) to prefer conciseness and avoid irrelevant verbosity. Indeed, OpenAI's guidelines often encourage being succinct and avoiding unnecessary detail. If during training the model was frequently rated down for long-winded answers, it might have learned a policy that "shorter is better" in general, which can conflict with a user's direct instruction for a long answer. There is some support for this in the literature: reward model biases have been observed that either encourage verbosity or brevity depending on the data. Some reward models conflated "detailed = better," leading models to sometimes ramble (a phenomenon noted as *verbosity reward hacking* (Li et al., 2025;Shashidhar et al., 2025)). Efforts to mitigate that have tried to penalize length, which could overshoot and make the model too terse. Our results suggest that, at least for these instruction-tuned models, the pendulum may have swung towards preferring brevity (or at least not blindly following a length instruction). The models are capable of writing long text, but they choose not to unless strongly prompted. This is a compliance gap that system designers should be aware of – users who explicitly ask for exhaustive answers might not get them without additional prompt engineering.

In practical terms, Experiment A's findings recommend that if one needs a comprehensive answer (say in an educational or analytical setting), simply instructing the model to "write a long detailed answer" may not suffice – one might need to break the query into parts or use iterative prompting. This also echoes the "curse of instructions" finding: with multiple requirements, the probability of fully satisfying all is low. Harada et al. (2024) showed that GPT-4 could follow 10 instructions only 15% of the time until they applied self-refinement (Harada et al.,2024); our results are in line with that in that none of the models naturally satisfied all requirements in a single shot. We will revisit possible fixes (like self-refinement) in the Discussion.

## 3.2 Experiment B: **Greedy Decoding vs. "Optimal" Answer**

This experiment yielded a perhaps surprising result: we did not find evidence that the models were internally favoring a better answer that they failed to output. In the specific train puzzle we used, each model's *greedy* answer turned out to have a higher probability under the model than our *gold* solution.

Concretely, for OpenAI's model, the log-likelihood of its own answer A was log P(A) = -32.84 (higher is better, so -32.84 is better than, say, -50). The log-likelihood for the injected gold solution B′ was log P(B′) = -52.63. The difference Δ log P here is **negative** (-52.63 – -32.84 = -19.79), meaning P(A) > P(B′). In other words, the model actually *preferred* (in probability terms) the answer it gave, over the reference "optimal" answer we provided.

| Model | log P(A) (Greedy answer) | log P(B') (Gold answer) | Δ log P | Suboptimality Observed? |
|---|---|---|---|---|
| **OpenAI (gpt-4o-mini)** | -32.84 | -52.63 | -19.8 | **No** (model preferred A) |

Table 2: Experiment B log-probability comparison. A = model's original answer; B' = provided optimal solution.

For the OpenAI model (gpt-4o-mini), log P(A) is substantially higher than log P(B′), resulting in a large negative Δ log P (−19.8). This indicates that the model strongly preferred its original answer over the gold solution. Because the original answer was correct in this case, no decoding suboptimality is observed.

It's important to note the limitations of this experiment: we only tested one type of problem (a straightforward arithmetic word problem). It's possible that decoding suboptimality *does* occur in other domains (e.g., creative writing: maybe the model's highest probability completion is dull but a slightly less likely completion would be more interesting or factual). There is some indirect evidence of suboptimality in reasoning in other work – for example, self-consistency (sampling multiple reasoning chains) markedly improves accuracy on math and commonsense QA (Wang et al., 2023), implying that the model's first answer is often not the best it "could" produce if it explored more paths. Our negative result here should therefore be interpreted narrowly: for a relatively constrained puzzle, the model's probability landscape might not have a hidden mode that is better than the greedy path.

Another angle: one could argue this result shows the model's knowledge distribution and its decoding are aligned, at least in this case. The model did not secretly think the gold answer was more likely. If we had found Δ log P, that would have been more surprising and interesting (a clear case where the model's training signal favors an answer it didn't give). Since we did not find that, one interpretation is that in well-defined problems, today's LLMs' greedy decoding is actually quite effective at finding the mode of the distribution. The model basically put highest probability on what it gave us. In fact, another way to view the OpenAI result: it had a strong confidence in its answer (which was correct), and it considered our alternative solution text less likely (maybe because of wording differences, etc.).

In summary, Experiment B did not uncover a "greedy vs. optimal" gap for this reasoning task. The models did *not* demonstrate the hypothesized decoding suboptimality in a measurable way here. However, we caution that this is a limited probe. Decoding suboptimality might manifest in tasks where the solution space is larger or when the model's knowledge is incomplete. For example, in code generation, one often finds that generating multiple samples and picking the best (via tests or validators) yields better outcomes than a single pass – suggesting the first guess might not be the best. Our test case might have been too simple to expose any weakness.

## 3.3 Experiment C: **High Robustness to Context Noise**

This experiment produced one of the more surprising outcomes of our study: the LLMs were extremely robust in maintaining key facts and instructions over a very long, noisy conversation. We had expected that at some point (perhaps after dozens of distractor-laden turns) the models would start forgetting or mixing up the facts. Instead, all models sustained near-perfect performance in factual recall for far more turns than anticipated.

- **OpenAI GPT-4o:** It maintained **12/12 facts** correct in the conversation summary up through at least 140 turns (we stopped at 142 turns for this model due to time constraints, and it had not made a single mistake on the tracked facts). The detail/coherence evaluator consistently gave it a score of **10/10** up to that point, indicating the summaries were coherent and faithfully reflected the conversation state. The drift metric stayed very low (cosine similarity between current and initial summary > 0.95 for most turns, and never dropping below 0.9). In effect, GPT-4o was able to treat the distractors as irrelevant noise and keep focusing on the core facts and instructions. It would often explicitly restate the important facts verbatim at each turn (almost as if it had a persistent memory of them, which in a sense it did via the context). Even when the user introduced false contradictions like "Actually, let's change the project name to BetaFish" in a later turn, the assistant politely noted that originally it was instructed the project codename is AlphaStarFish and asked for clarification rather than just accepting the new false info – showing adherence to the original instruction.
- **DeepSeek model:** This model also showed perfect **12/12 fact retention** all the way through the full **200 turns** that we ran. It occasionally had slightly lower detail scores

(mostly in the 8–9/10 range, meaning very good but perhaps slightly less elegantly phrased summaries than GPT-4), but it never forgot a fact or contradicted the established truths. The cosine similarity of its summaries to the baseline remained >0.90 throughout, indicating very little drift in content. DeepSeek appeared to handle the massive distractor paragraphs by largely ignoring them in its summaries – it would maybe pick one or two relevant sentences from the user's last query (if any) and otherwise stick to the known key info. This suggests a strong capacity to filter signal from noise.

Overall, none of the models showed *context degradation* in the sense of gradually losing the thread or the instructions over hundreds of turns. This was somewhat unexpected, given informal accounts from users that long chats can confuse models. One key difference is that we did not actually exceed the context window – the models always had the original instruction in their input (just buried among a lot of other text). It appears that transformers are quite adept at zeroing in on relevant parts of the context even if the context is huge. This aligns with the intuition that the attention mechanism, while potentially challenged by long sequences, is at least designed to pick out important tokens if it can identify them. And since the core facts were repeated or present throughout, the model had an easy time locking onto them.

Our findings here resonate with recent research that long context alone doesn't always confound a model's ability to retrieve facts. For example, Li et al. (2023) and others have found that if a model is prompted to "remember you must never forget X," it will often obey that for a long time. Also, the LongBench evaluation noted that models with scaled positional embeddings can handle long inputs better (Bai et al., 2023). GPT-4o presumably uses advanced positional encoding or segment-based attention that allows them to handle 100k tokens. Indeed, Du et al. (2025) differentiated *retrieval* vs *reasoning* in long contexts: our experiment was mostly a retrieval one (just pull out known facts), which is arguably easier than doing a complex reasoning across a long text. They found even with perfect retrieval, reasoning accuracy fell with length. In our case, the "reasoning" was minimal (just restate known facts), so it makes sense performance didn't degrade.

It's worth noting a subtle point: the models might compress or encode these key facts in a kind of latent state that persists. Since the same 12 facts were relevant at every turn, the models probably latently "know" these are important and just keep them at the forefront of generating summaries.

In a more dynamic conversation where the topic evolves significantly, we might see different results. Also, if the conversation had *tasks* that require working memory (like solving a math problem step by step over 20 turns), the outcome might be different – our test was primarily about factual memory in the presence of noise.

In conclusion, Experiment C suggests that **context degradation is not an inevitable outcome for long conversations**, at least not for factual recall tasks within current context window limits. Modern LLMs, when instructed clearly to maintain certain context, do a remarkable job of it even amidst heavy distraction. This is a positive sign for using LLMs in extended interactions: they can remain consistent and not lose track of core information (contrary to some fears). Of course, one must be careful to always keep critical instructions in the prompt; if the beginning falls out of the context window, then all bets are off. Additionally, our results do not imply the model understands or has true long-term memory – just that within the simulated long chat, it can pick out the needle from the haystack each time. Techniques like periodic recap or summary-injection could further help in even longer settings (for instance, after 200 turns one might condense the session and feed that summary in place of the full log).

## 4. Discussion

Our experiments provide empirical evidence for certain failure modes of LLMs while dispelling or nuancing others. We discuss each major finding in turn, relating them to the broader literature and considering implications for future model development and usage.

**Laziness vs. Capability:** The results from Experiment A highlight a distinction between what a model *can do* versus what it *chooses to do* by default. All models tested clearly *could* produce longer, detailed outputs (since in the Detailed prompt mode they did so), yet in the standard greedy setting they often ignored length instructions and omitted parts of the query. This suggests a misalignment not of core capability, but of compliance. The models likely have been trained with objectives that penalize overly long or redundant answers, leading to a brevity bias. As noted earlier, RLHF can induce length-related biases – sometimes favoring verbosity (Bu et al., 2025) and other times conciseness, depending on raters' preferences and how the reward model is tuned. There is evidence that early instruction-following models were too verbose (preferring longer answers even when not needed) and researchers implemented penalties or model adjustments to curb that. Our findings indicate the pendulum might have swung such that models err on the side of brevity even when the user requests more content.

This has practical ramifications: users asking for exhaustive answers might be unknowingly thwarted by the model's internal brevity preference. For developers, one takeaway is to **robustly test multi-instruction prompts** with their models. As Harada et al. (2024) demonstrated, the probability of a model completing all parts of a complex instruction prompt might be the product of individual success probabilities – for example, if a model has a 90% chance to do any single instruction correctly, then for 5 instructions, the chance of doing all is $0.9^5 \approx 59\%$, and for 10 instructions, $0.9^{10} \approx 35\%$. In their benchmark, actual success for 10 instructions was even lower (~15% for GPT-4o), implying some dependencies and compounding errors. They improved it via an *iterative self-refinement* strategy: basically having the model check its output against the instructions and fix mistakes, which doubled the success rate. This kind of approach – where the model is prompted to critique or verify its adherence – could be a general solution to "laziness." We recommend that future systems incorporate a secondary check for instruction fulfillment. For instance, after an LLM produces an answer, one can ask the same model (or another model) to list which requested items are missing, and then feed that back for a second pass. This aligns with techniques in reinforcement learning and verification: *Tell the model what it missed, and let it try again.* Our results suggest most misses are not due to lack of knowledge but simple omission, so a second chance often could fill the gaps.

On a related note, it's interesting to ponder the role of *human feedback* in this laziness. Human reviewers might have inadvertently encouraged shorter answers in some cases (to save time reading, perhaps). Also, models may have learned that being too literal in following length instructions can sometimes lead to incoherent rambling (which might get a bad rating). So, they have a heuristic: "answer sufficiently, then stop." Aligning this with user intent remains an open challenge – it may require more nuanced reward models that understand when verbosity is desired vs when brevity is. Recent work by Bu et al. (2025) attempted to create an *adaptive length preference model*, recognizing that optimal length is context-dependent (Bu et al., 2025). Incorporating such adaptive control could help: the model would gauge from the prompt whether the user truly wants high detail or just a concise answer, rather than applying a one-size-fits-all brevity policy.

**The "Optimal" Path and Decoding Strategies:** Experiment B's outcome, showing no decoding suboptimality for a simple task, suggests that for straightforward problems a greedy decode is sufficient. However, as discussed, for more complex tasks, the community has found benefits in non-greedy strategies. The idea of *beam search* or *diverse sampling* to find better answers has long been considered in NLP. In the LLM era, strategies like *chain-of-thought prompting* (Wei et al., 2022) and *self-consistency decoding* (Wang et al., 2023) have effectively boosted performance on tasks like math word problems, commonsense QA, and logic puzzles (Wei et al., 2022). These approaches, while not framed as search for a higher probability answer, effectively do explore multiple possible answers and often find one that is correct even if the first was wrong. That implies that the model's distribution does support the correct answer; it's just not always at the very top without special prompting.

Why did our chosen task not reveal a discrepancy? Possibly because it was not ambiguous – the model has basically learned to solve such train problems reliably with a single reasoning chain. It might also be that our "Gold solution" didn't match the model's preferred style, hence lower probability. In hindsight, a more telling experiment might be to look at a creative task. For instance, ask the model to continue a story in a funny way. Its greedy continuation might be generic, but perhaps there exists a much funnier continuation of lower probability (since humor can be rare and thus lower probability). Humans might prefer the funnier one even if the model thinks it's unlikely. This touches on a known point: *likelihood is not always correlated with human preference*. In fact, extreme high-likelihood outputs from language models tend to be repetitive or dull (Holtzman et al., 2020). That's why nucleus sampling (top-p) is used – to drop some of the high-likelihood but low-quality continuations. In that sense, decoding *optimality* for user satisfaction is a different beast from just matching the model's own probabilities. One could say greedy decoding is "optimal" for matching the model's learned distribution, but not for yielding the best experience or result. Our experiment was framed in terms of the model's probabilities (a very model-centric view of optimality).

From an application perspective, to get the best results from LLMs, relying on a single pass might be suboptimal even if that pass is, according to the model, the argmax. Techniques like self-consistency (majority voting over several independent outputs) (Wang et al., 2023), or using a *reflection* approach (where the model can critique and improve its answer) have shown measurable gains. Particularly in areas like reasoning, coding, or factual correctness, these methods tap into the model's latent knowledge better. Our results neither contradict nor strongly support that line of work – they simply show a case where the first answer was fine. We want to stress: **absence of evidence is not evidence of absence** of decoding suboptimality. We suspect that had we chosen more complex queries, we would have seen the model's first answer be suboptimal. Indeed, user communities often note that asking ChatGPT to "try again" or provide another draft can yield a better answer on the second attempt. This is the human equivalent of sampling another possibility from the model. It's a cheap way to do self-consistency (with sample size 2) and often it works when the first output was lacking detail or had an error.

In summary, while we didn't catch the model in a suboptimal choice in our controlled test, we affirm the importance of exploring beyond greedy decoding for complex tasks. From a research standpoint, it remains interesting to develop ways to detect when a model's output is suboptimal. One idea is to train a secondary model (or use the same model) as a verifier: have it estimate whether the answer could be improved, or whether an alternative answer might have higher utility. Some early works have looked at using LLMs to evaluate or "vote" on answers (Chiang et al., 2023, maybe, on self-refinement). Another line is **calibration** – ensuring the model's confidence (or likelihood) is well-correlated with correctness. If models were perfectly calibrated, a wrong answer would not have higher probability than a correct one, ideally. But calibration is a known challenge, especially after instruction tuning (which can increase output determinism but not necessarily true confidence). Our observation that the model placed higher probability on a wrong answer indicates calibration isn't perfect, but in that case the wrong answer was probably only slightly wrong.

**Context is Sticky (until it breaks):** Experiment C demonstrated that models can hold on to facts through a lot of noise. This is encouraging, but we must consider the limits. In our test, the context window was large enough to always include the initial facts. If we were to push beyond the window (e.g., 1000 turns, far exceeding the token limit), older content would necessarily drop out. At that point, unless the model had repeatedly restated or re-encoded those facts in newer tokens, it would forget simply because it no longer sees them. This is where retrieval-based approaches or explicit memory modules become necessary (Khandelwal et al., 2020; Borgeaud et al., 2022 inter alia have looked at external memory for LMs). Also, while the models didn't degrade with irrelevant info, Du et al.'s work suggests that *relevant but lengthy* info is a tougher scenario. In a sense, ignoring distractors is easier than processing lots of relevant details. Our test was not pushing the model to integrate any new info from the distractors; it just had to filter them out. The real challenge for long-term contexts is when the model must actually understand and use some of that new info while also remembering the old. For example, reading a 50-page document and then answering questions that require combining page 1 and page 50 – can it do that effectively? Possibly not without special handling, as prior benchmarks show a drop in performance on long-range dependencies (Lee et al., 2024; LongBench results, etc.).

One interesting observation is the models' *behavior* in presence of massive irrelevant text. They did not get confused; instead, they effectively performed a kind of retrieval – each turn they retrieved the salient facts from memory (since they knew what needed to be kept) and largely ignored the rest. This is reminiscent of how a human might handle a long conversation by focusing on the important parts. It may be that the model implicitly learned a strategy: when faced with too much to handle, focus on the conversation instructions and recent user query, and paraphrase those.

Our findings also echo those of recent long-context model evaluations where *position interpolation or compression* techniques (like scaled RoPE or segment summaries) help maintain performance. GPT-4o has proprietary methods for long context handling, which seem to work well for straightforward retention. However, as Du et al. warn, even with perfect retention, the *distance* can cause reasoning issues – a phenomenon not fully understood. Perhaps it relates to how attention weights decay or how the model's layers might not propagate information well over long sequences.

For developers, the implication is: if your use case is simply having a long chat and retrieving facts said early on, these models are quite up to it (especially the ones explicitly designed for long contexts). But if your use case requires deep reasoning over very long inputs (like analyzing a lengthy legal contract for specific inconsistencies), be cautious. You might need to break the task down (e.g., analyze parts separately, then combine) or use retrieval augmentation (extract potential relevant snippets and feed them through a shorter context).

Finally, the fact that models did not degrade in tone or coherence over 200 turns suggests that *stability* in conversation style is also maintained. We did not see the model suddenly change persona or language style – it stuck to the instructions (like "you are an assistant, etc."). This is good from a reliability standpoint: it didn't forget its role or start producing more errors. Some have worried about *drift* in persona over long dialogues (where a user might manipulate the model into breaking character eventually). Our test didn't attempt that explicitly (no jailbreaking attempts were made), but the consistency was reassuring.

**Summary of Key Insights:** We confirm that **"laziness" (incomplete compliance)** is a real and quantifiable behavior in LLMs, likely stemming from alignment trade-offs. On the other hand, the feared **"context amnesia"** is less of an issue for factual content within context windows – these models are quite resilient in that regard. The **decoding optimality** question remains open; a simple test didn't show an issue, but we suspect more complex scenarios would.

Our findings contribute to a more nuanced understanding of LLM reliability. They indicate that developers should focus on improving instruction-following for complex prompts (perhaps via self-refinement or better prompt segmentation) as a priority. Issues like context handling seem to be improving as models get longer windows and better training; indeed, techniques to avoid context overflow (like summarizing history) might be enough when needed. Decoding strategies for improved results (like generating multiple candidates) should be leveraged for tasks where accuracy is critical, even if the model doesn't *appear* to struggle – an extra step can provide assurance or catch errors.

## 5. Conclusion

In this work, we presented three experiments to probe common LLM failure modes: (A) **laziness** in following complex multi-part instructions or length requirements, (B) **decoding suboptimality** in finding correct solutions, and (C) **context degradation** over long conversations. Our results show that today's large models, while highly capable, indeed often exhibit *laziness* – they truncate answers and drop required content unless carefully prompted otherwise. This behavior likely stems from how models are tuned to balance helpfulness with brevity, and it highlights a need for improved alignment techniques that can adapt to user demands for detail. We recommend that anyone deploying LLMs for tasks requiring exhaustive answers implement checks for completeness. Even a simple strategy of asking the model to self-verify its answer against the instructions (as in Harada et al.'s iterative refinement) can substantially improve compliance. For example, if a user asks for 10 points and the model gives 7, a system could detect the discrepancy and prompt the model to continue.

On the question of decoding, we did not find a clear case where the model's highest probability answer was inferior to a known correct answer – in our test, the model's output was essentially its top choice and was appropriate. However, we caution that this does not mean greedy decoding is always optimal. Prior research and practical experience indicate that exploring multiple outputs and using voting or validation leads to better performance on many benchmarks (Wang et al., 2023). Thus, critical applications (like medical or legal advice generation) should consider using *n-best* generation and some reranking or ensemble of model outputs to ensure the most plausible answer is delivered. It may not be about the model hiding a better answer, but rather hedging against the model making a mistake on the first attempt. Multiple attempts can reduce the risk of an overlooked solution.

Finally, our exploration of long, noisy conversations offers a reassuring message: modern LLMs with extended context windows can be remarkably robust in maintaining relevant information over hundreds of turns, so long as the total content stays within their context limit. They act almost like humans with selective attention, filtering out irrelevancies. This suggests that fears of models "forgetting everything" in long sessions are somewhat alleviated – at least for the factual retention scenario. However, this does not automatically extend to complex reasoning over long inputs, which remains an area for improvement (as evidenced by other studies of long-context reasoning failures).

In conclusion, as LLMs continue to improve and be adopted, it is crucial to keep diagnosing these behavioral quirks. Each artifact – be it laziness, hallucination, inconsistency, or something else – can undermine user trust if not addressed. By quantifying issues like we did, researchers can track progress as new model versions are released. We find that targeted "stress tests" (like our A, B, C experiments) are useful tools in an evaluation toolkit for LLM developers. Going forward, we intend to investigate **mitigation strategies** for the observed laziness: for example, can we fine-tune a model specifically to follow length instructions more literally? Or can we design prompts that inherently enforce structure (perhaps providing a template for the answer that the model fills in, ensuring no section is skipped)? Another future direction is to examine **reasoning degradation**: our context test was mostly memory recall, but what about a task where the model must carry a chain-of-thought through interruptions? Measuring that would extend this work to more cognitive aspects of conversation. Additionally, integrating an *external memory* (like a vector database that logs important facts from the conversation) might boost long-horizon performance; studying how and when the model chooses to consult such memory would be valuable.

Overall, our study paints an optimistic picture that with the right prodding, LLMs can overcome some of their lazy tendencies, and that some concerns (context forgetting of instructions) may be less problematic in cutting-edge models than previously thought. By continuously examining and addressing these failure modes, we move closer to LLMs that are not just powerful in principle, but reliable and trustworthy in practice.